\documentclass[letterpaper]{article} 
\usepackage[]{aaai24}  
\usepackage{times}  
\usepackage{helvet}  
\usepackage{courier}  
\usepackage[hyphens]{url}  
\usepackage{graphicx} 
\usepackage{natbib}  
\usepackage{caption} 
\usepackage{algorithm}
\usepackage{algorithmic}
\usepackage{amsmath}
\usepackage{amssymb}
\usepackage{multirow}
\usepackage{makecell}
\usepackage{threeparttable}

\usepackage{newfloat}
\usepackage{listings}
\DeclareCaptionStyle{ruled}{labelfont=normalfont,labelsep=colon,strut=off} 
\floatstyle{ruled}
\newfloat{listing}{tb}{lst}{}
\floatname{listing}{Listing}
\usepackage{color}

\usepackage[switch]{lineno}

\title{Approximated Prompt Tuning for Vision-Language Pre-trained Models}
\author{
Qiong Wu$^{12}$, Shubin Huang$^{1}$, Yiyi Zhou$^{12}$, Pingyang Dai$^{1}$, Annan Shu$^{3}$, Guannan Jiang$^{3}$, Rongrong Ji$^{12}$ \\
$^{1}$ Key Laboratory of Multimedia Trusted Perception and Efficient Computing, \\
Ministry of Education of China, Xiamen University, 361005, P.R. China. \\
$^{2}$ Institute of Artificial Intelligence, Xiamen University, 361005, P.R. China. \\
$^{3}$ Intelligent Manufacturing Department, Contemporary Amperex Technology Co. Limited \\
{\tt\small \{qiong, shubinhuang\}\protect\@stu.xmu.edu.cn, \{zhouyiyi, pydai\}\protect\@xmu.edu.cn,} \\
{\tt\small \{shuan01, jianggn\}\protect\@catl.com, rrji\protect\@xmu.edu.cn}
}
\usepackage{bibentry}

\begin{document}

\maketitle

\begin{abstract}
Prompt tuning is a parameter-efficient way to deploy large-scale pre-trained models to downstream tasks by adding task-specific tokens. 
In terms of vision-language pre-trained (VLP) models, prompt tuning often requires a large number of learnable tokens to bridge the gap between the pre-training and downstream tasks, which greatly exacerbates the already high computational overhead.
In this paper, we revisit the principle of prompt tuning for Transformer-based VLP models, and reveal that the impact of soft prompt tokens can be actually approximated via independent information diffusion steps, thereby avoiding the expensive global attention modeling and reducing the computational complexity to a large extent.
Based on this finding, we propose a novel \emph{Approximated Prompt Tuning} (APT) approach towards efficient VL transfer learning. 
To validate APT, we apply it to two representative VLP models, namely ViLT and METER, and conduct extensive experiments on a bunch of downstream tasks.
Meanwhile, the generalization of APT is also validated on CLIP for image classification and StableDiffusion for text-to-image generation.
The experimental results not only show the superior performance gains and computation efficiency of APT against the conventional prompt tuning methods, \emph{e.g.}, $+7.01\%$ accuracy and $-82.30\%$ additional computation overhead on METER, but also confirm its merits over other parameter-efficient transfer learning approaches\footnote{Our code is given in supplementary materials and will be publicly released after acceptance.}.
\end{abstract}

\section{Introduction}

%
Prompt tuning~\cite{conf/acl/LiL20, conf/acl/CuiWLYZ21, conf/icml/RadfordKHRGASAM21, journals/corr/abs-2103-10385, journals/corr/abs-2203-12119, conf/cvpr/ZhouYL022, journals/ijcv/ZhouYLL22} is a parameter-efficient way to adapt large-scale pre-trained models to downstream tasks.
It inserts multiple prompt tokens into the input sequence to unify the pre-trained and downstream data distributions~\cite{conf/emnlp/PetroniRRLBWM19, conf/icml/RadfordKHRGASAM21}, thereby avoiding the expensive full fine-tune of pre-trained models.
Recent advances~\cite{conf/acl/LiL20, journals/corr/abs-2203-12119, conf/cvpr/ZhouYL022, journals/ijcv/ZhouYLL22} resort to trainable tokens to replace the hand-craft ones for the adaption on downstream tasks, which is named \emph{soft prompt tuning}~\cite{conf/acl/LiL20, journals/corr/abs-2103-10385}.
In terms of the way of insertion, soft prompt tuning can be further categorized into \emph{deep prompt tuning}~\cite{journals/corr/abs-2203-12119} and \emph{shallow prompt tuning}~\cite{conf/acl/LiL20}, respectively.
Currently, prompt tuning has achieved great success in \emph{natural language processiong} (NLP)~\cite{conf/emnlp/PetroniRRLBWM19, conf/acl/LiL20, conf/emnlp/LesterAC21} and also been recently applied to shallow-fusion based vision-language pre-trained (VLP) models like CLIP~\cite{conf/icml/RadfordKHRGASAM21} for image classification~\cite{conf/cvpr/ZhouYL022, journals/ijcv/ZhouYLL22}.

\begin{figure}[t]
\centering
\includegraphics[width=1.0\columnwidth]{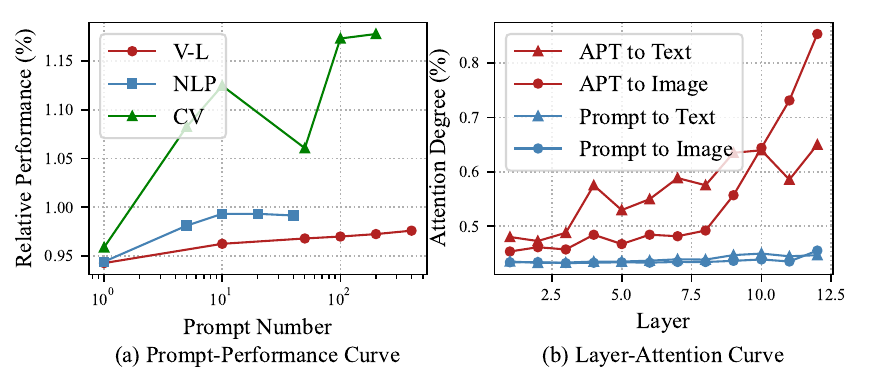}
\caption{
\small
(a) The relative performance of soft prompt tuning to full fine-tune \protect\emph{w.r.t.} the prompt number on VL, NLP and CV tasks~\protect\footnotemark. VL tasks often require more prompt tokens than single-modal ones. 
(b) The comparison between our APT and soft prompt tuning in terms of information diffusion efficiency, \protect\emph{i.e.}, the attention weights to the input sequence. 
The adaption efficiency of soft prompting is much less efficient on VL tasks.
}
\label{fig1:Motivation}
\end{figure}
\footnotetext{These results are from VPT~\cite{journals/corr/abs-2203-12119} on VTAB datasets (CV), Prefix~\cite{conf/acl/LiL20} on DART (NLP) and deep prompting on VQA (V-L) }

However, it is still more challenging to apply prompt tuning to common VLP models, \emph{e.g.}, ViLT~\cite{conf/icml/KimSK21} and METER \cite{conf/cvpr/DouXGWWWZZYP0022},  than the language ones.
Above all, most VL tasks are greatly different from the VL pre-training objectives, \emph{e.g.},such as \emph{visual question answering} (VQA)~\cite{conf/cvpr/GoyalKSBP17}.
Thus, hand-craft prompts often fail to adapt to these tasks.
Soft prompt tuning is applicable to VLP models~\cite{journals/corr/abs-2203-12119}.
But due to the large task gap and the increase of modalities, more learnable tokens are often required to adapt VLP models to downstream tasks, as shown in Fig.~\ref{fig1:Motivation}-a.
In addition, we also notice that even with a bunch of tokens, soft prompt tuning still has limited impacts on the input sequence, \emph{i.e.}, attention weights, leading to a sub-optimal adaption, as shown in Fig.~\ref{fig1:Motivation}-b.  
Considering that these tokens are often involved in self-attention, of which computation is quadratic to the input length~\cite{conf/nips/VaswaniSPUJGKP17}, this inefficient adaption will significantly increase the already high computational overhead of VLP models.

\begin{figure}[t]
\centering
\includegraphics[width=1.0\columnwidth]{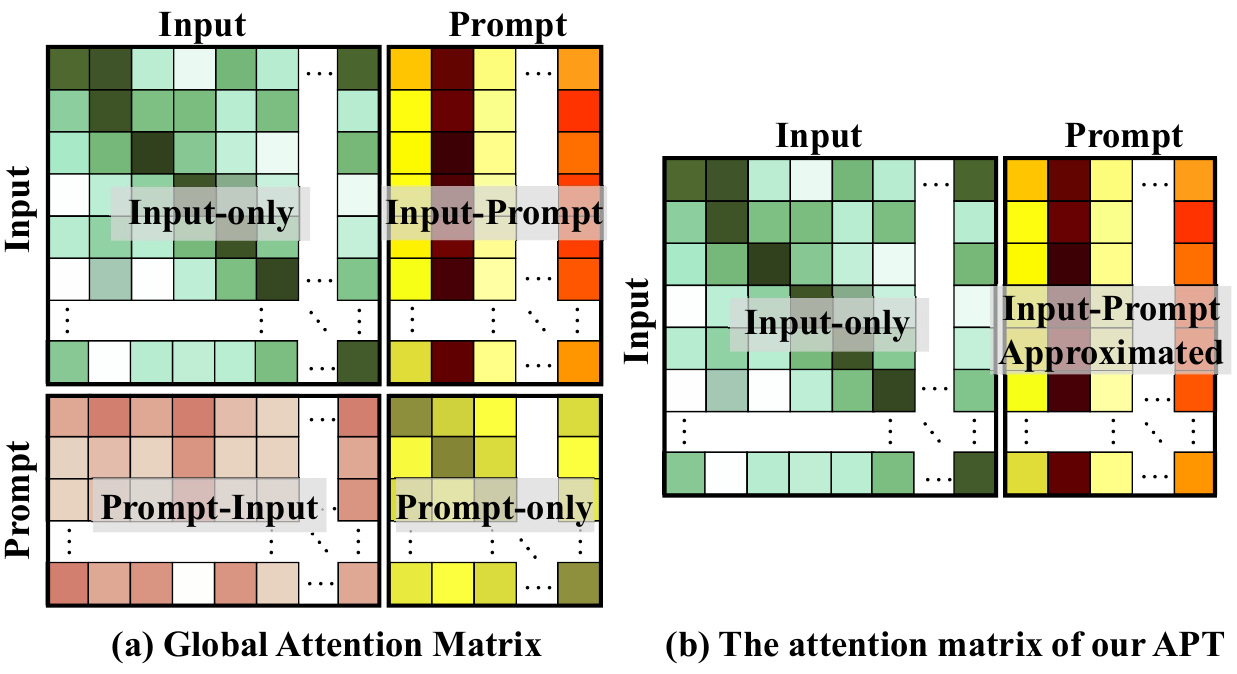}
\caption{\small
%
%
%
The illustrations of global self-attention matrices with and without APT. (a) is the attention matrices of common prompt tuning. In (b), the prompt-input and prompt-only parts are removed, and the input-prompt attention is approximated by APT. 
}
\label{fig1.1:subparts}
\end{figure}

By revisiting the principle of prompt tuning, we find that there exists a potential solution for efficient VL adaption.
Particularly, prompt tuning aims to use additional tokens to influence the input sequence, so as to minimize the gap between pre-training and downstream tasks~\cite{conf/emnlp/PetroniRRLBWM19, conf/acl/CuiWLYZ21}. 
In terms of soft prompt tuning, the tokens are usually inserted into the self-attention layers of VLP models~\cite{conf/acl/LiL20, journals/corr/abs-2203-12119, conf/cvpr/ZhouYL022}.
Via analyzing self-attention, we observe that the obtained attention weight matrix can be actually divided into four sub-parts as shown in Fig.~\ref{fig1.1:subparts}-a. 
Here, we call them \emph{input-only}, \emph{input2prompt}, \emph{prompt2input} and \emph{prompt-only} attention matrices, respectively.
Under the setting of deep prompt tuning~\cite{journals/corr/abs-2203-12119}, \emph{i.e.}, the prompts are layer-wise,  the computations of \emph{prompt2input} and \emph{prompt-only} can be indeed skipped and will not affect the prompt tuning of next layer. 
And the \emph{input-only} is the default operation of pre-trained models that cannot be changed. 
In this case, the key to improving prompt tuning lies on the \emph{input2prompt}, which is essentially an information diffusion step from the prompt tokens to the input sequence under the perspective of graph theory~\cite{conf/mm/ZhouJSLHSD020}. 
However, we find that its functionality can be actually approximated via a more effective process independent to global attention, thereby improving the efficiency of VL adaption.

Motivated by this observation, we propose a novel \emph{approximated prompt tuning} (APT) approach for VLP models in this paper.
Similar to deep prompt tuning~\cite{journals/corr/abs-2203-12119, conf/acl/LiL20}, APT inserts a set of learnable tokens into each self-attention layer of the VLP model.
As shown in Fig.~\ref{fig1.1:subparts}-b, a key difference is that we separate these tokens from the expensive global self-attention and approximate their effects independently by aggregating the prompt tokens with low-rank transformations.
In this way, the proposed APT can effectively diffuse more information from prompt tokens to the input sequence while avoiding the expensive global self-attention, as shown in Fig.~\ref{fig1:Motivation}-b. 
%

%
To validate APT, we apply it to two deep-fusion based VLP models, namely ViLT~\cite{conf/icml/KimSK21} and METER~\cite{conf/cvpr/DouXGWWWZZYP0022}, on three VL benchmarks including VQA~\cite{conf/iccv/AntolALMBZP15}, NLVR$^2$~\cite{conf/acl/SuhrZZZBA19} and Flickr30K~\cite{journals/ijcv/PlummerWCCHL17}.
In addition, we also examine its generalization on CLIP~\cite{conf/icml/RadfordKHRGASAM21} for the \emph{base-to-new} classification task~\cite{journals/ijcv/ZhouYLL22} and on StableDiffusion~\cite{conf/cvpr/RombachBLEO22, ruiz2022dreambooth} for text-to-image generation.
The experimental results well confirm the obvious merits of APT over the conventional prompt tunning methods~\cite{conf/acl/LiL20, journals/corr/abs-2203-12119, conf/cvpr/ZhouYL022, journals/ijcv/ZhouYLL22}, \emph{e.g.}, $+8.30\%$ accuracy on VQA2.0 for METER while reducing up to $17.70\%$ additional computation overhead. Our APT also yield better performance than most parameter-efficient transfer learning (PETL) approaches~\cite{journals/corr/abs-2203-12119, conf/iclr/HuSWALWWC22, conf/cvpr/Sung0B22, conf/iclr/HeZMBN22}, \emph{e.g.} $70.94\%$ on VQA2.0 for ViLT and $80.97\%$ on NLVR$^2$ for METER. 
%

Overall, our contributions are three-fold:
\begin{itemize}
\item We identify the key challenges of prompt tuning on common VLP models, \emph{e.g.} ViLT~\cite{conf/icml/KimSK21} and METER~\cite{conf/cvpr/DouXGWWWZZYP0022}, which are excessive computation overhead and low prompt tuning efficiency.
\item We propose a novel \emph{approximated prompt tuning} (APT) method for both parameter- and computation-efficient prompt tuning, which approximates the influence of prompt tokens via independent aggregation steps. 
\item The proposed APT not only outperforms existing prompt tuning methods but also achieves better performance than other PETL approaches on 2 VLP models and 4 VL tasks.
Its generalization is also validated on CLIP and StableDiffusion. 
\end{itemize}

\section{Related Work}

\subsection{Vision-Language Pre-training}

%
Similar to NLP pre-training paradigms~\cite{conf/naacl/DevlinCLT19, conf/nips/BrownMRSKDNSSAA20, journals/jmlr/RaffelSRLNMZLL20, conf/acl/LewisLGGMLSZ20, journals/corr/abs-1907-11692, conf/iclr/LanCGGSS20}, vision-language (VL) pre-training also apply generative prediction tasks, \emph{e.g.} \emph{masked language modeling} (MLM) and \emph{maksed image modeling} (MIM), to achieve self-supervised learning on massive cross-modal data. 
A key difference is that VLP models usually require two encoders to process image and language information, \emph{e.g.}, BERT~\cite{conf/naacl/DevlinCLT19} and Faster-RCNN~\cite{conf/nips/RenHGS15}, and their Transformer-based backbone networks are not only used for high-level representation learning, but also for cross-modal deep fusion and interaction~\cite{journals/corr/abs-1908-03557, conf/iclr/SuZCLLWD20, conf/eccv/ChenLYK0G0020, journals/corr/abs-2004-00849, journals/corr/abs-1907-11692, conf/icml/KimSK21, conf/cvpr/DouXGWWWZZYP0022, conf/cvpr/HuangZH0FF21}. 
METER~\cite{conf/cvpr/DouXGWWWZZYP0022}, the base model of this paper, is the typcial model using this paradigm. Meanwhile, we also validate APT on the other representative model called ViLT~\cite{conf/icml/KimSK21}, which process the image and text information with only one end-to-end Transformer network. 

\subsection{Prompt Tuning}

Prompt tuning~\cite{conf/nips/BrownMRSKDNSSAA20, conf/emnlp/PetroniRRLBWM19, conf/acl/LiL20, conf/acl/CuiWLYZ21, conf/icml/RadfordKHRGASAM21, journals/corr/abs-2203-12119, conf/cvpr/ZhouYL022, journals/ijcv/ZhouYLL22, journals/corr/abs-2103-10385} is a parameter-efficient way to adapt pre-trained models to downstream tasks.
Concretely, for hand-crafted prompts~\cite{conf/emnlp/PetroniRRLBWM19, conf/icml/RadfordKHRGASAM21}, it often inserts a pre-defined prompt phrase into the input sequences, thus reminding the model of pre-trained knowledge, \emph{e.g.}, ``\emph{This is a picture of} [X]''.
However, hard prompt tuning heavily relies on manual design.
To overcome this issue, soft prompt tuning~\cite{conf/acl/LiL20, journals/corr/abs-2203-12119, journals/ijcv/ZhouYLL22} is proposed to automatically learn trainable prompts via downstream task adaption. 
%
In terms of the prompt placement, soft prompt tuning can be further divided into two patterns, \emph{i.e.}, the shallow~\cite{conf/acl/LiL20} and the deep~\cite{journals/corr/abs-2203-12119} ones.  
Shallow prompt tuning methods~\cite{conf/emnlp/LesterAC21, conf/acl/LiL20} only expand the input sequence with trainable vectors at the first layer, while deep prompt tuning methods~\cite{journals/corr/abs-2203-12119} expand the input sequence between any two layers with trainable tokens.
%
%

\subsection{Parameter Efficient Transfer Learning}

Parameter Efficient Transfer Learning (PETL)~\cite{conf/icml/HoulsbyGJMLGAG19, conf/acl/MahabadiR0H20, conf/eccv/ZhangSZGM20, conf/acl/GuoRK20, conf/nips/SungNR21, conf/nips/MahabadiHR21, conf/cvpr/Sung0B22, conf/iclr/HuSWALWWC22,  conf/iclr/HeZMBN22, conf/acl/MaoMHAM0YK22} aims to update a small number of parameters to approach the fully-tuned performance on downstream tasks. 
%
%
%
In addition to prompt tuning, a common paradigm is the adapter-based methods \cite{conf/icml/HoulsbyGJMLGAG19, conf/acl/MahabadiR0H20, conf/nips/MahabadiHR21, journals/corr/abs-2110-04544, journals/corr/abs-2111-03930, conf/cvpr/Sung0B22}, called \emph{Adapter} for short, which insert lightweight networks into the pre-trained model to project hidden features onto downstream data space.
%
To avoid the additional computation overhead during inference, Hu \emph{et al.} propose a low-rank adaption (LoRA)~\cite{conf/iclr/HuSWALWWC22} method, based on weight re-parameterization. 
%
%
%
%
In the field of vision-language learning, VL-adapter~\cite{conf/cvpr/Sung0B22} insert low-dimensional networks into a pre-trained language model to adapt to common VL tasks~\cite{journals/corr/ChenFLVGDZ15, conf/cvpr/GoyalKSBP17, conf/acl/SuhrZZZBA19}.
Its key difference to this paper is that the language model is not VL pre-trained, lacking enough generalization for common VLP models. 

\section{Preliminary}

Before introducing our approach, we first recap the principle of prompt tuning for VLP models.
Concretely, given a pre-trained vision-language (VLP) model, denoted as $G(\cdot)$, and the image-text example of the downstream task, denoted as $(I,T)$, the target of prompt tuning is to minimize the adaption loss with a set of prompt tokens $\mathbf{P} \in \mathbb{R}^{p \times d}$:
\begin{equation}
\begin{aligned}
\operatorname*{argmin}_{\mathbf{P}} \mathcal{L}\big(G(I,T,P|\theta^+)\big),
\end{aligned}
\label{opt_objective}
\end{equation}
where $\theta^+$ is the pre-trained weights of $G$ and will be fixed during prompt tuning\footnote{In most case, the classifier will be trained for a specific task}.
$\mathcal{L}$ is the objective function of the downstream task.

Considering that the parameters are fixed during adaption, the features of the input sequence are hard to update for the downstream task. 
In this case, prompt tokens $\mathbf{P}$ are often used in the self-attention of VLP models for diffusing task-related information to the input sequence $\mathbf{X} \in \mathbb{R}^{n \times d}$:
\begin{equation}
\begin{aligned}
[\mathbf{X}' || \mathbf{P}'] &= SA(\mathbf{X} || \mathbf{P}), \\
\end{aligned}
\label{}
\end{equation}
where $SA(\cdot)$ represents the self-attention module.
$\mathbf{X}'$ and $\mathbf{P}'$ are the corresponding outputs of $\mathbf{X}$ and $\mathbf{P}$, respectively.
In particular, X' and P' are obtained by
\begin{equation}
\begin{aligned}
\mathbf{X}' &= \mathbf{A}_{I}\mathbf{X}\mathbf{W}_v + \mathbf{A}_{IP}\mathbf{P}\mathbf{W}_v, \\
\mathbf{P}' &= \mathbf{A}_{PI}\mathbf{X}\mathbf{W}_v + \mathbf{A}_{P}\mathbf{P}\mathbf{W}_v, 
\end{aligned}
\label{tot_attention}
\end{equation}
where $\mathbf{A}_{I}$, $\mathbf{A}_{IP}$, $\mathbf{A}_{PI}$ and $\mathbf{A}_{P}$ are the sub-attention matrices, corresponding to the \emph{input-only}, \emph{input2prompt}, \emph{prompt2input} and \emph{prompt-only} parts described in introduction and shown in Fig.~\ref{fig1.1:subparts}.
$\mathbf{W}_q$, $\mathbf{W}_k$ and $\mathbf{W}_v$ are the weight matrices of $Q$, $K$, $V$ projections in $SA$.
Under the layer-wise setting~\cite{journals/corr/abs-2203-12119}, the prompt tokens are initialized for each layer and will not be used in the next $SA$. In this case, the computation of $\mathbf{P}'$ can be indeed removed, which can reduce the complexity by $O(2pd^2+4npd+2p^2d)$, where $p$ is often a large value on VL tasks.
%

%
Eventually, the feature update of VLP models with prompt tokens can be relaxed to  
\begin{equation}
\begin{aligned}
\mathbf{X}' &= \mathbf{A}_{I}\mathbf{X}\mathbf{W}_v + \mathbf{A}_{IP}\mathbf{P}\mathbf{W}_v\\
            &= \frac{\mathbf{\gamma}_{I}}{\mathbf{\gamma}_{I} + \mathbf{\gamma}_{IP}} \sigma(\frac{\mathbf{X}\mathbf{W}_q(\mathbf{X}\mathbf{W}_k)^T}{\sqrt{d}})\mathbf{X}\mathbf{W}_v \\
            &+ \frac{\mathbf{\gamma}_{IP}}{\mathbf{\gamma}_{I} + \mathbf{\gamma}_{IP}} \sigma(\frac{\mathbf{X}\mathbf{W}_q(\mathbf{P}\mathbf{W}_k)^T}{\sqrt{d}})\mathbf{P}\mathbf{W}_v, \\
\text{where \ } & \mathbf{\gamma}_{I} = \sum e^{\mathbf{Q}\mathbf{K}^T_i}, \mathbf{\gamma}_{IP} = \sum e^{\mathbf{Q}{\mathbf{P}_k}^T_j}
\end{aligned}
\label{direct_calculate}
\end{equation}
Here, $\sigma(\cdot)$ is the Softmax function, and $\mathbf{\gamma}_{I}$ and $\mathbf{\gamma}_{IP}$ are the attention proportion for the input sequence and prompt tokens, respectively.
In Eq.~\ref{direct_calculate}, the first term is the self-attention update of input features, which is the compulsory operation of VLP models. 
To this end, the effectiveness of prompt tuning lies in the second term, which is essentially a weighted information diffusion step from $\mathbf{P}$ to $\mathbf{X}$.
Since the scale-dot product is still required, this diffusion step is also expensive.

\section{Approximated Prompt Tuning}

Based on the above observation, we propose \emph{approximated prompt tuning} (APT) to model the attention impacts of prompt tokens. 
In particular, we can get the prompt tuning process of APT with the following formula:
\begin{equation}
\begin{aligned}
\mathbf{X}'= SA(\mathbf{X}) + APT(\mathbf{X}, \mathbf{P}).
\end{aligned}
\label{target}
\end{equation}
For simplicity, we regard the information diffusion from P to X as $\Delta X$:
\begin{equation}
\begin{aligned}
\Delta \mathbf{X} &= APT(\mathbf{X}, \mathbf{P}) \\
&= \frac{\mathbf{\gamma}_{IP}}{\mathbf{\gamma}_{I} + \mathbf{\gamma}_{IP}} \sigma(\mathbf{X}\mathbf{W}_q(\mathbf{P}\mathbf{W}_k)^T)\mathbf{P}\mathbf{W}_v.  \\
\end{aligned}
\label{delta_item}
\end{equation}
%
%

%
To approximate $\Delta X$, we first focus on the information aggregation of prompt tokens, denoted as $\Delta X'$:
\begin{equation}
\begin{aligned}
\Delta \mathbf{X}' = \sigma(\mathbf{X}\mathbf{W}_q\mathbf{W}_k^T\mathbf{P}^T)\mathbf{P}\mathbf{W}_v.
\end{aligned}
\label{approximated_delta_item_without_weight}
\end{equation}
Note that, $\mathbf{W}_v$ is fixed in SA, and $\mathbf{P}$ is a trainable matrix. 
To this end, we can directly update the projection of prompt tokens onto the $V$ subspace, \emph{i.e.}, put $\mathbf{P}\mathbf{W}_v$ as $\mathbf{P}'$.
Similarly, we can simplify $\mathbf{P}\mathbf{W}_k\mathbf{W}_q^T$ as $K$.
Then, the $\textbf{X}$ can be directly taken as the $Q$ without projection, and the computation from transforming $\mathbf{X}$ and $\mathbf{P}$ into $Q$, $K$ and $V$ of $SA$ can be saved.

%
Next, we show that $V$ can be linearly transformed to $K$:
\begin{equation}
\begin{aligned}
\Delta \mathbf{P} & = \mathbf{P}\mathbf{W}_k\mathbf{W}_q^T - \mathbf{P}\mathbf{W}_v, \\
                & = \mathbf{P}(\mathbf{W}_k\mathbf{W}_q^T - \mathbf{W}_v).
\end{aligned}
\label{difference_between_KV}
\end{equation}
%
%
Here, $\Delta \mathbf{P}$ denotes the difference between $V$ and $K$.
Because $V$ can be transformed to $K$ by a linear transformation, we approximate Eq.~\ref{approximated_delta_item_without_weight} as
\begin{equation}
\begin{aligned}
\Delta \mathbf{X}' = \sigma\big(\mathbf{X}(\mathbf{P}'\mathbf{W}_p + \mathbf{P}')^T\big)\mathbf{P}',
\end{aligned}
\label{approximated_delta_item_without_weight_and_qkv}
\end{equation}
where $\mathbf{W}_p \in \mathbb{R}^{d \times d}$ aims at transforming prompt tokens from $V$ to $K$.
However, calculating $\mathbf{P}\mathbf{W}_p$ is still not cheap due to the high feature dimension.

As the low intrinsic dimension component~\cite{conf/iclr/LiFLY18, conf/acl/AghajanyanGZ20} plays a dominant role in model optimization, the rank for $\mathbf{W}_p$ is finite according to the theorem of the rank of matrices:
\begin{equation}
\begin{aligned}
rank(\mathbf{W}_p) \leq rank(\mathbf{W}_k\mathbf{W}_q^T) + rank(\mathbf{W}_v),
\end{aligned}
\label{rank_theorem}
\end{equation}
where $rank(\cdot)$ is the rank of the matrix.
%
We can approximate the aggregation of prompt tokens in a low-rank way:
\begin{equation}
\begin{aligned}
\Delta \mathbf{X}' = \sigma\big(\mathbf{X}(\mathbf{P}'\mathbf{W}_1\mathbf{W}_2 + \mathbf{P}')^T\big)\mathbf{P}'.
\end{aligned}
\label{approximated_P_K}
\end{equation}
Here, $\mathbf{W}_1 \in \mathbb{R}^{d \times r}$ and $\mathbf{W}_2 \in \mathbf{R}^{r \times d}$ are two low-dimensional matrix, where $r \ll d$.
The rank of projection matrix $\mathbf{W}_1\mathbf{W}_2$ is limited by $r$.
%
%
%
The way we obtain $Q$, $K$ and $V$ matrices for attention modeling is cheaper than the original global attention.

\begin{table*}[t]
\caption{
Comparisons of APT and the conventional prompt tuning methods for ViLT and METER on VQA, NLVR$^2$ and Flickr30K.
The best performance is \textbf{bold} while the second one is \underline{underlined}.
}
\centering
\resizebox{0.95\textwidth}{!}
{
\tiny
\begin{tabular}{c | l | cc | c c cc | c}
\hline
\multirow{2}{*}{\textbf{Backbone}} & \multirow{2}{*}{\textbf{Method}} & \multirow{2}{*}{\makecell[c]{\textbf{Updated} \\ \textbf{Parameter}}} & \multirow{2}{*}{\makecell[c]{\textbf{Additional} \\ \textbf{FLOPs}}} & \textbf{VQA} & \textbf{NLVR$^2$} & \multicolumn{2}{c|}{\textbf{Flickr30K}} &  \multirow{2}{*}{\textbf{Avg.}} \\
& & & & test-dev & test-P & IR R@1 & TR R@1 \\
\hline
\multirow{4}{*}{ViLT}
& Full Tuning        & 115.43M & 0.0 & 71.26 & 76.13 & 64.40 & 83.50 & 73.82 \\
& Shallow Prompt     & 0.15M & 19.53G & 66.47 & 66.47 & 55.92 & 74.80 & 65.92 \\
& Deep Prompt        & 1.84M & 5.14G & \underline{69.30} & \underline{73.34} & \underline{58.64} & \underline{79.50} & \underline{70.20} \\
\cline{2-9}
& \textbf{APT}       & 1.92M & 0.91G & \textbf{70.94} & \textbf{75.92} & \textbf{63.26} & \textbf{81.60} & \textbf{72.93} \\
\hline
\multirow{4}{*}{METER} 
& Full Tuning        & 323.31M & 0.0 & 77.43 & 83.05 & 82.22 & 94.30 & 84.25 \\
& Deep Prompt        & 3.68M & 13.05G & 67.57 & \underline{65.79} & 70.90 & 87.70 & 72.99 \\
& Shallow Prompt     & 0.30M & 28.71G & \underline{68.51} & 65.69 & \underline{74.20} & \underline{88.60} & \underline{74.25} \\
\cline{2-9}
& \textbf{APT}       & 3.83M & 2.31G  & \textbf{75.45} & \textbf{80.97} & \textbf{80.88} & \textbf{92.90} & \textbf{82.55} \\
\hline
\end{tabular}
}
\label{table_comparison_ViLT}
\vspace{2mm}
\end{table*}

\begin{figure*}[t]
\centering
\includegraphics[width=1.0\textwidth]{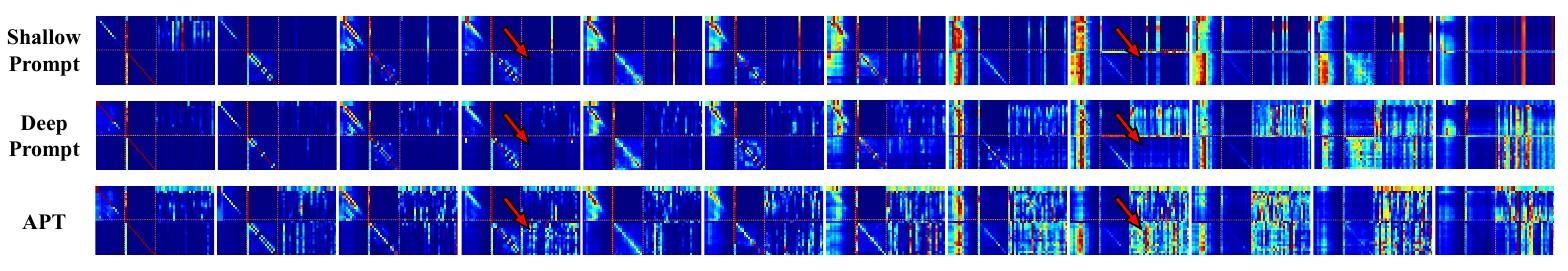}
\caption{ \small
The visualizations of the attention results of shallow prompt, deep prompt and our APT with ViLT on VQA2.0 dataset.
The color denotes the degree of attention, while the redder the higher and \emph{vice versa}.
Compared with shallow prompt and deep prompt, APT can more effectively diffuse prompt information to the input sequence from the low layers of ViLT, see the red arrows.
}
\label{fig3:Visualization}
\end{figure*}

Afterwards, we consider the way to merge the original output of the self-attention module $SA(\mathbf{X})$ and the information of prompt tokens $\Delta \mathbf{X}$.
Because the calculation of attention is still related to the input sequence, it is difficult to reduce the complexity of the approximation via independent computation.
In this case, to adaptively adopt the impact of each prompt token, a simple solution is activating the attention matrix with $ReLU$ instead of $Softmax$ and omitting the weight item.
Then, Eq.~\ref{delta_item} can be represented as
\begin{equation}
\begin{aligned}
\Delta \mathbf{X} = \psi\big(\mathbf{X}(\mathbf{P}'\mathbf{W}_1\mathbf{W}_2 + \mathbf{P}')^T\big)\mathbf{P}',
\end{aligned}
\label{approximated_delta_item}
\end{equation}
where $\psi(\cdot)$ represent ReLU activation.
In this manner, the weights for prompts depend on the norm of prompt tokens and their relation to input sequence.

Furthermore, from the weight calculation in Eq.~\ref{delta_item}, we observe that the effect of prompt tokens is not only influenced by their dependency to the input sequence, but also by the sum of attentions to the input sequence.
With the intrinsic of the Softmax function that the maximum value has the most impact, we re-define Eq.\ref{delta_item} by 
\begin{equation}
\begin{aligned}
\Delta \mathbf{X} &= \alpha \cdot \psi(\mathbf{X}(\mathbf{P}'\mathbf{W}_1\mathbf{W}_2 + \mathbf{P}')^T)\mathbf{P}', \\ 
\text{where} \alpha &= max\{ \mathbf{P}'\mathbf{W}_1\mathbf{W}_2 + \mathbf{P}'\},
\end{aligned}
\label{approximated_delta_item_with_max}
\end{equation}
%
where $max \{ \cdot \}$ is the maximum function for the weight of each token.
Thus, APT can globally adjust the information diffusion from the prompt tokens.
Since the activation function of Eq.~\ref{approximated_delta_item_with_max} no longer relies on the original attention matrix, the APT is easier to deploy for VLP models.


Up to now, we have fully considered the effect of prompt tokens in  diffusing task-related information to the input sequence.
Then, we also take into account the effect of prompt tokens on the original attention matrix.
As shown in Eq.~\ref{direct_calculate}, the information diffusion also influences the original attention matrix by increasing the denominator of the weight for the item from VLP module.
To this end, we add a learnable scale $s$ for the entire output, and the proposed APT can be summarised as follow:
\begin{equation}
\begin{aligned}
\mathbf{X}' &= \mathbf{A}_{I}\mathbf{X}\mathbf{W}_v + \mathbf{A}_{IP}\mathbf{P}\mathbf{W}_v \\
            &\approx e^s \cdot \big( SA(\mathbf{X}) 
            + \alpha \cdot \psi(\mathbf{X}(\mathbf{P}'\mathbf{W}_1\mathbf{W}_2 + \mathbf{P}')^T)\mathbf{P}' \big), \\ 
& \text{where} \ \alpha = max\{ \mathbf{P}'\mathbf{W}_1\mathbf{W}_2 + \mathbf{P}' \}.
\end{aligned}
\label{summary}
\end{equation}
Here, the learnable value $s$ control the total amount of information diffused by APT and also make the output of attention modules fit the following layers.

Eventually, the proposed APT method separates the effect of prompt tokens from the original attention module.
The independence of APT brings two main benefits:
(1) Information diffusion can break the limitation of patterns from VLP model, \emph{i.e.} not constrained by Softmax-based normalization.
(2) The computational overhead is significantly reduced by about $O(2pd^2)$. In practice, it can save about $82.30\%$ and $62.62\%$ computations for ViLT~\cite{conf/icml/KimSK21} and METER~\cite{conf/cvpr/DouXGWWWZZYP0022} compare to conventional prompt tuning methods.

%
\section{Experiments}

\subsection{Datasets and Experimental Setup}

%
%
\textbf{Dataset and Metric.} VQA2.0~\cite{conf/cvpr/GoyalKSBP17} is one of the most popular datasets for visual question answering (VQA) task.
It uses images from MS-COCO~\cite{ren2015exploring} and has about $443,757$, $214,254$ and $447,793$ VQA examples for training, validation and testing, respectively.
%
%
%
NLVR$^2$~\cite{conf/acl/SuhrZZZBA19} is built for visual reasoning.
It contains $107,292$ examples of human-written English sentences for pairs of photographs.
%
%
Flickr30k~\cite{journals/ijcv/PlummerWCCHL17} is a widely-used benchmark dataset in this image-text matching task.
The dataset consists of $31,783$ images, and each has five corresponding captions.
For CLIP, we validate APT on $11$ popular image classification datasets, including ImageNet~\cite{conf/cvpr/DengDSLL009}, Caltech101~\cite{journals/cviu/Fei-FeiFP07}, OxfordPets~\cite{conf/cvpr/ParkhiVZJ12}, StandfordCars~\cite{conf/iccvw/Krause0DF13}, Flowers102~\cite{conf/icvgip/NilsbackZ08}, Food101~\cite{conf/eccv/BossardGG14}, FGVCAircraft~\cite{journals/corr/MajiRKBV13}, SUN397~\cite{conf/cvpr/XiaoHEOT10}, DTD~\cite{conf/cvpr/CimpoiMKMV14}, EuroSAT~\cite{journals/staeors/HelberBDB19}, UCF101~\cite{journals/corr/abs-1212-0402}~\footnote{The details of these datasets are given in Appendix.}.
This comprehensive benchmark comprises datasets that cover a diverse set of vision tasks, including classification on generic objects, scenes, actions, and fine-grained categories. It also includes specialized tasks like recognizing textures and satellite imagery.

\noindent \textbf{Implementation details.} 
%
%
We validate APT on two deep-fusion based VLP models, namely ViLT~\cite{conf/icml/KimSK21} and METER~\cite{conf/cvpr/DouXGWWWZZYP0022}, and one shallow-fusion based VLP network called CLIP~\cite{conf/icml/RadfordKHRGASAM21}. 
%
%
In terms of ViLT, we add APT to its each SA layer.  
We set the rank value $r$ in Eq.~\ref{approximated_P_K} to $4$ and the number of prompt tokens $p=200$ as the default setting.
The prompt tokens are initialized by a normal distribution with a mean of $0.0$ and a variance of $0.02$.
And we only apply a single attention rather than the multi-head one~\cite{conf/naacl/DevlinCLT19} for the proposed APT method.
During the training, we update the classifier, class tokens and modal-type embeddings, while the rest parameters of ViLT are kept fixed.
For each task, we follow its default settings and increase the learning rate by five times.
In terms of METER, APT is inserted into its self-attention and cross-attention layers. 
The rest settings are the same as ViLT. 
For CLIP~\cite{conf/icml/RadfordKHRGASAM21}, we insert APT into the self-attention layers of its text encoder, and we set the rank $r=2$ and the number of prompts $p=4$.
APT is optimized by SGD with a learning rate of $2 \times 10^{-4}$ and weight decay of $0.3$ for $10$ epochs.
%
%
Following~\cite{conf/icml/RadfordKHRGASAM21}, we also use a hard prompt phrase of ``\emph{a photo of} [X]'', which is fed to the text encoder of CLIP.

\subsection{Experimental results}

\begin{table*}[t]
\caption{
Comparisons of APT and the state-of-the-art PETL methods for ViLT and METER on VQA, NLVR$^2$ and Flickr30K.
%
%
The best performance is \textbf{bold} and the second best is \underline{underlined}.
}
\centering
\resizebox{0.95\textwidth}{!}
{
\tiny
\begin{tabular}{c | l | cc | c c cc | c}
\hline
\multirow{2}{*}{\textbf{Backbone}} & \multirow{2}{*}{\textbf{Method}} & \multirow{2}{*}{\makecell[c]{\textbf{Updated} \\ \textbf{Parameter}}} & \multirow{2}{*}{\makecell[c]{\textbf{Additional} \\ \textbf{FLOPs}}} & \textbf{VQA} & \textbf{NLVR$^2$} & \multicolumn{2}{c|}{\textbf{Flickr30K}} &  \multirow{2}{*}{\textbf{Avg.}} \\
& & & & test-dev & test-P & IR R@1 & TR R@1 \\
\hline
\multirow{8}{*}{ViLT}
& Full Tuning        & 115.43M & 0.0 & 71.26 & 76.13 & 64.40 & 83.50 & 73.82 \\
& Classifier Only    & -       & 0.0 & 65.75 & 66.08 & 57.42 & 78.00 & 66.81 \\
\cline{2-9}
& Deep Prompt        & 1.84M & 5.14G & 69.30 & 73.34 & 58.64 & 79.50 & 70.20 \\
& LoRA               & 0.15M & 0.0 & 68.44 & 72.77 & 57.44 & 77.70 & 69.09 \\
& Scaled PA          & 1.80M & 0.44G & 70.40 & 75.13 & 61.88 & 79.00 & 71.60 \\
& Adapter            & 3.56M & 0.86G & \underline{70.85} & \underline{75.51} & \underline{62.68} & \underline{81.40} & \underline{72.61} \\
\cline{2-9}
& \textbf{APT}       & 1.92M & 0.91G & \textbf{70.94} & \textbf{75.92} & \textbf{63.26} & \textbf{81.60} & \textbf{72.93} \\
\hline
\multirow{7}{*}{METER}
& Full Tuning        & 323.31M & 0.0 & 77.43 & 83.05 & 82.22 & 94.30 & 84.25 \\
& Classifier Only    & -     & 0.0 & 69.93 & 73.23 & 78.80 & 89.00 & 77.74 \\
\cline{2-9}
& Deep Prompt        & 3.68M & 13.05G & 67.57 & 65.79 & 70.90 & 87.70 & 72.99 \\
& LoRA               & 0.29M & 0.0  & 74.00 & 78.82 & 79.86 & 92.60 & 81.32 \\
& Adapter            & 5.34M & 1.64G  & 74.70 & 79.93 & 80.38 & 91.90 & 81.73 \\
& Scaled PA          & 3.82M & 1.12G  & \underline{75.36} & \underline{79.86} & \underline{80.30} & \underline{91.80} & \underline{81.83} \\
\cline{2-9}
& \textbf{APT}       & 3.83M & 2.31G  & \textbf{75.45} & \textbf{80.97} & \textbf{80.88} & \textbf{92.90} & \textbf{82.55} \\
\hline
\end{tabular}
}
\label{table_comparison_METER}
\vspace{4mm}
\end{table*}

\noindent\textbf{Comparison with prompt tuning methods.} 
We first compare APT with two common soft prompt tuning methods, \emph{i.e.}, deep prompt~\cite{journals/corr/abs-2203-12119} and shallow prompt~\cite{conf/acl/LiL20}, in Tab.~\ref{table_comparison_ViLT}. For all methods, the number of prompts is set to $200$ for a fair comparison. 
%
%
%
From Tab.~\ref{table_comparison_ViLT}, the performance of existing prompt tuning methods is far behind the full tuning one, \emph{i.e.} $-7.90\%$ to $-3.62\%$ on ViLT and $-11.26\%$ to $-10.00\%$ on METER.
These results are also worse than their performance on NLP~\cite{conf/acl/LiL20} and vision tasks~\cite{journals/corr/abs-2203-12119}, showing the challenge of prompt tuning on VLP models.
Among these compared methods, Deep Prompt shows better results than shallow prompts at most cases, while its parameter size is larger and similar to APT.
Notably, the average improvements of APT to these prompt methods are $+2.73\%$ to $+7.01\%$ on ViLT and $+8.30\%$ to $+9.56\%$ on METER, respectively, while the saved additional computations can be up to $82.30\%$ on ViLT and $91.95\%$ on METER. 
Meanwhile, the performance of APT almost approaches full tuning, \emph{e.g.}, $-0.89\%$ and $-1.70\%$ on average for ViLT and METER, respectively. 
Considering the small number of parameters updated, these results are indeed significant.
%
\begin{figure}[t]
\centering
\includegraphics[width=0.85\columnwidth]{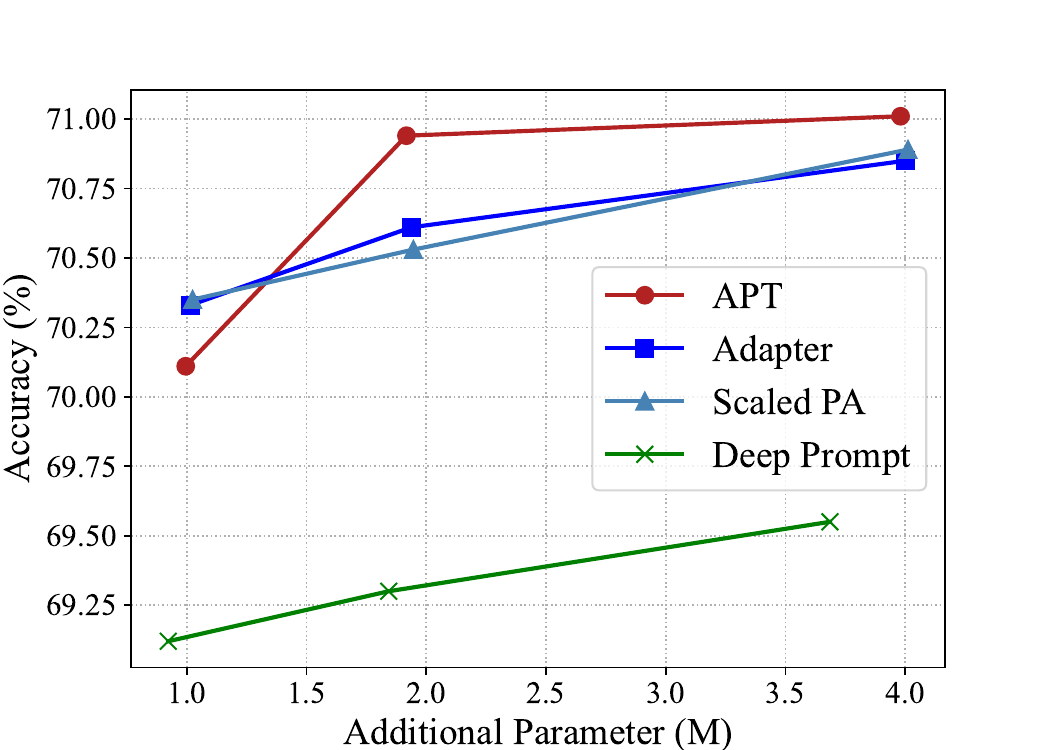}
\caption{ \small
The comparison between APT and other PETL methods in terms of performance and parameter size. 
APT has a better trade-off between performance and parameter costs. 
}
\label{fig4:PPcurve}
\end{figure}

To obtain more intuitive comparisons, we also visualize the attentions of these prompt tuning methods in Fig.~\ref{fig3:Visualization}. 
In this figure, We select the 15 most active tokens of the visual and text inputs, and the top 30 prompt tokens for visualization, 60 tokens in total. 
The global attention matrices can be divided into six sub-parts, \emph{i,e.} \emph{Text-Text}, \emph{Text-Image}, \emph{Text-Prompt}, \emph{Image-Text}, \emph{Image-Image}, \emph{Image-Prompt}.
From these examples, we can first observe that in the lower layers of the VLP model, the information exchanges mainly happens among the tokens of the same modality, and the prompts barely affect the input sequence. 
As the inference progress, their impacts of common prompts become slightly more pronouns. In terms of shallow prompts, the impact of its tokens is still marginal, while deep prompt will be better at the last few layers of the model.
The above results are also consistent with their performance on VL tasks.
In stark contrast, APT can effectively diffuse prompt information to the input sequence of the VLP models, see the arrows. 
And its attention weights become more intensive in the higher layers, suggesting its effectiveness towards task adaption.

\noindent \textbf{Comparison with existing PETL methods.}
Next, we compare APT with a bunch of PETL methods, including LoRA~\cite{conf/iclr/HuSWALWWC22}, VL-Adapter (Adapter)~\cite{conf/cvpr/Sung0B22} and Scaled Parallel Adapter (Scaled PA)~\cite{conf/iclr/HeZMBN22}, of which results are given in Tab.~\ref{table_comparison_METER}\footnote{These results are reproduced by us because there are no ready-made literature to refer. Details are given in Appendix.}. 
%
%
From this table, we can first see that LoRA is most efficient in both parameters and computation due to its low-rank re-parameterization scheme.
Compared to pre-trained language models~\cite{journals/corr/abs-1907-11692, conf/nips/BrownMRSKDNSSAA20}, its performance on VLP models is much inferior, especially on the tasks that are greatly different from pre-training, \emph{e.g.}, VQA and NLVR$^2$, suggesting the challenge of VL adaption. 
We can also find that although the adapter-based methods show better adaptabilities than LoRA, they still perform worse than our APT. 
Compared with VL-Adapter, APT can achieve obvious gains on ViLT and METER, while saving about $46.07\%$ and $28.28\%$ parameters, respectively.
In terms of the most advanced Scaled PA, APT is slightly inferior in parameter and computation costs, but its adaption performance is consistently better than Scaled PA on two VLP models.
Overall, these results suggest that our APT is a competitive method in PETL with great potential.  

%
In Fig.~\ref{fig4:PPcurve}, we also present the performance comparison of APT to other PETL methods with different parameter costs.
It can be seen that Deep Prompt are much inferior than other methods in terms of parameter efficiency and performance, suggesting its difficulty on VL adaption.
Adapter~\cite{conf/cvpr/Sung0B22} and Scaled PA~\cite{conf/iclr/HeZMBN22}, as the advanced PETL methods, are all parameter efficient, and their adaptions are also plausible on VQA. However, the overall performance of these two methods is close, which is beyond expectation. 
Compared to these adapter-based methods, the performance of APT can achieve obvious gains at a scale of about 2M parameters, which becomes stable as the parameter size grows.

\noindent \textbf{Ablation Study.} We first examine the impact of prompt number and the rank value in Eq.~\ref{approximated_P_K}, of which results are given in Tab.~\ref{table_weight_ablation}. 
Here, ``\emph{identify}'' denotes that directly using prompt tokens as $K$ and $V$ in Eq. \ref{approximated_delta_item_without_weight_and_qkv}, while ``\emph{dense}'' means that low-rank transformation is not used in Eq.~\ref{approximated_P_K}. 
The first observation from Tab.~\ref{table_weight_ablation} is that the increase of prompt tokens is beneficial to VLP models, which can obtain improvements on both tasks, \emph{e.g.}, 100 to 200.
However, when exceeding 200, its gains are marginal in contrast to other prompt methods as shown in Fig.~\ref{fig4:PPcurve}, which also suggests the effectiveness of APT for VLP models.
In terms of the rank value, the performance of ``identity'' suggests that directly using prompt tokens for attention is suboptimal.
And the low-rank approximation can better trade off performance and parameter cost, \emph{e.g.} rank value $r=4$, which is even superior than the dense transformation on NLVR$^2$.  
Overall, these results well confirm the effectiveness of APT towards efficient VL adaption. 
More experiments can refer to our \textbf{Appendix}. 


\noindent \textbf{Generalization on CLIP.} 
\begin{table}[t]
\caption{
Ablation study on different constructions and the number of prompt tokens.
%
%
*the default setting.
}
\centering
\resizebox{0.45\textwidth}{!}
{
\begin{tabular}{c | c | c | c c}
\hline
\multirow{2}{*}{{Prompt}} & \multirow{2}{*}{{Rank value}} & \multirow{2}{*}{\makecell[c]{{Additional} \\ {Parameter}}} & {VQA} &  {NLVR$^2$} \\    
& & & test-dev & test-P \\
\hline
200   & Identity                          & 1.84M & 70.42 & 75.27 \\
200   & dense               & 8.93M & 71.07 & 75.67 \\
\hline
100   & 4 & 0.99M & 70.11 & 74.81 \\
150   & 4 & 1.46M & 70.49 & 74.87 \\
\hline
400   & 4 & 3.76M & 70.64 & 75.68 \\
400   & 16 & 3.98M & 71.01 & 75.95 \\
\hline
200*   & 4* & 1.92M & 70.94 & 75.92  \\
\hline
\end{tabular}
}
\vspace{2mm}
\label{table_weight_ablation}
\end{table}
\begin{table}[t]
\caption{
Comparison of zero-shot CLIP (\emph{CLIP}), CoOp, CoCoOp and APT on the \emph{base to new} classification task.
%
}
\centering
\tabcolsep=0.7mm
\resizebox{0.48\textwidth}{!}
{
\footnotesize
\begin{tabular}{l | c c c c | c c c c}
\hline
\multirow{3}{*}{Dataset} & \multicolumn{8}{c}{Method} \\ 
\cline{2-9}
& \multicolumn{4}{c |}{Base} & \multicolumn{4}{c}{New} \\
\cline{2-9}
& CLIP & CoOp & CoCoOp & APT & CLIP & CoOp & CoCoOp & APT \\
\hline
ImgNet
& 72.43 & \textbf{76.47} & \underline{75.98} & 75.97 & 68.14 & 67.88 & \underline{70.43} & \textbf{71.23} \\
Cal101
& 96.84 & \textbf{98.00} & \underline{97.96} & 97.93 & \underline{94.00} & 89.81 & 93.81 & \textbf{94.13} \\
Pets
& 91.17 & 93.67 & \textbf{95.20} & \underline{94.97} & 97.26 & 95.29 & \textbf{97.69} & \underline{97.60} \\
Cars
& 63.37 & \textbf{78.12} & 70.49 & \underline{76.10} & \textbf{94.89} & 60.40 & 73.59 & \underline{75.13} \\
Flowers
& 72.08 & \textbf{97.60} & \underline{94.87} & 94.47 & \textbf{77.80} & 59.67 & \underline{71.75} & 70.13 \\
Food
& 90.10 & 88.33 & \textbf{90.70} & \underline{90.17} & \underline{91.22} & 82.26 & \textbf{91.29} & 90.70 \\
Aircraft
& 27.19 & \textbf{40.44} & 33.41 & \underline{38.63} & \textbf{36.29} & 22.30 & 23.71 & \underline{33.97} \\
SUN
& 69.36 & \underline{80.60} & 79.74 & \textbf{81.50} & 75.35 & 65.89 & \underline{76.86} & \textbf{78.20} \\
DTD
& 53.24 & \underline{79.44} & 77.01 & \textbf{81.00} & \textbf{59.90} & 41.18 & \underline{56.00} & 48.53 \\
SAT
& 56.48 & \textbf{92.19} & 87.49 & \underline{91.10} & \textbf{64.05} & 54.74 & 60.04 & \underline{62.13} \\
UCF
& 70.53 & \underline{84.69} & 82.33 & \textbf{85.13} & \underline{77.50} & 56.05 & \textbf{77.64} & 76.93 \\
\hline
Average                                  & 69.34 & \underline{82.69} & 80.47 & \textbf{82.72} & \textbf{74.22} & 63.22 & 71.69 & \underline{72.64} \\
\hline
\end{tabular}
}
\label{table_generalization}
\vspace{2mm}

\end{table}
We further examine the generalization ability of APT on the shallow-fusion based VLP model, \emph{i.e.}, CLIP~\cite{conf/icml/RadfordKHRGASAM21}, under the base-to-new classification task~\cite{conf/cvpr/ZhouYL022}, of which results are given in Tab.~\ref{table_generalization}. 
In this task, the model needs to adapt to the base dataset, and will be further evaluated on unseen data (new dataset).
The compared methods include zero-shot CLIP, CoOp~\cite{journals/ijcv/ZhouYLL22} and CoCoOp~\cite{conf/cvpr/ZhouYL022}.
The detailed settings are given in \textbf{Appendix}.
From this table, we first observe that zero-shot CLIP has a strong transferring learning ability. 
Due to its large-scale pre-training, it can obtain superior performance under the new task evaluations.
However, without tuning on base datasets, its performance is much inferior than PETL methods.
In terms of CoOp, it can achieve satisfactory performance for base task adaption.
However, its generalization is limited to new tasks, only $63.22\%$ on average, suggesting the over-fitting problem. 
In stark contrast, APT can obtain a good performance in adapting base tasks while generalizing well to new ones. 
Compared to latest PETL methods for CLIP, \emph{i.e.} CoCoOp, its performance is also consistently better under two settings.
These results confirm the generalization of APT. 

\noindent \textbf{Generalization on StableDiffusion.}
\begin{figure}[t]
\centering
\includegraphics[width=0.48\textwidth]{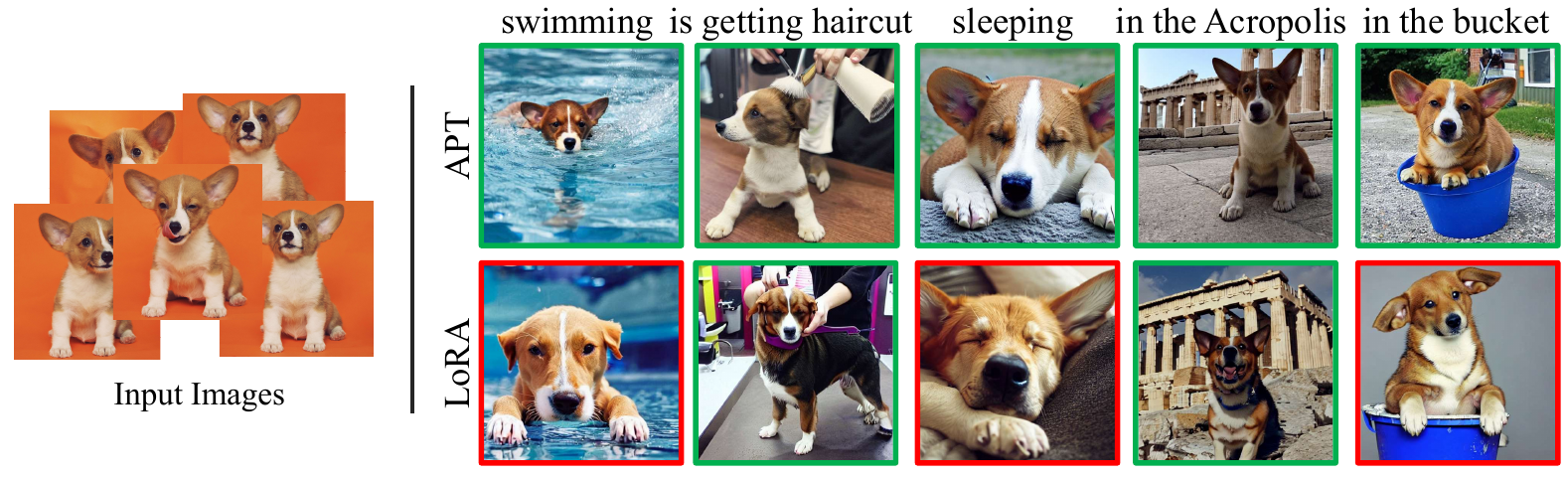}
\caption{ \small
The comparison between APT and LoRA on StableDiffusion under the setting of subject-driven image generation. 
The red boxes denote the failure cases. Compared with LoRA, APT can better customize the generation based on the reference images. 
}
\label{fig3:dif}
\vspace{2mm}
\end{figure}
We also examine the generalization ability of APT on the StableDiffusion~\cite{conf/cvpr/RombachBLEO22} following the setting of DreamBooth~\cite{ruiz2022dreambooth}, and the compared method is LoRA, of which results are given in Fig.~\ref{fig3:dif}. 
%
%
The detailed setting is given in Appendix.
From the image, we can find out that the dogs generated with different prompts can keep the same attribute.
Similar to LoRA, APT binds the attributes of the dog in the training set to specific vocabulary ``sks''.
These results suggest that APT is also capable of text-to-image generation.

\section{Conclusion}

In this paper, we focus on the issues of high computation overhead and inefficient adaption of prompt tuning on vision-language pre-trained (VLP) models.
By revisiting the principle of prompt tuning, we can figure out that the key to improve prompt tuning lies in its information diffusion to the input sequence, which can be indeed independent to the expensive global self-attention via effective approximations.
Motivated by this observation, we propose a novel \emph{approximation prompt tuning} (APT) approach towards effective VL adaption. 
APT approximates the impacts of prompt tokens to the input sequence via a low-rank token aggregation design, reducing the computation cost to a large extent. 
We validate APT on 2 VLP models and 3 VL benchmarks, and also generalize VPT to CLIP for image classification and StableDiffusion for subject-driven image generation. 
The quantitative and qualitative results not only show the obvious merits of APT over existing prompt tuning methods in both computation efficiency and performance, but also outperform the compared PETL methods on these VL tasks. 

\bibliography{aaai24}

\begin{thebibliography}{58}
\providecommand{\natexlab}[1]{#1}

\bibitem[{Aghajanyan, Gupta, and Zettlemoyer(2021)}]{conf/acl/AghajanyanGZ20}
Aghajanyan, A.; Gupta, S.; and Zettlemoyer, L. 2021.
\newblock Intrinsic Dimensionality Explains the Effectiveness of Language Model
  Fine-Tuning.
\newblock In Zong, C.; Xia, F.; Li, W.; and Navigli, R., eds., \emph{ACL},
  7319--7328.

\bibitem[{Antol et~al.(2015)Antol, Agrawal, Lu, Mitchell, Batra, Zitnick, and
  Parikh}]{conf/iccv/AntolALMBZP15}
Antol, S.; Agrawal, A.; Lu, J.; Mitchell, M.; Batra, D.; Zitnick, C.~L.; and
  Parikh, D. 2015.
\newblock {VQA:} Visual Question Answering.
\newblock In \emph{ICCV}, 2425--2433.

\bibitem[{Bossard, Guillaumin, and Gool(2014)}]{conf/eccv/BossardGG14}
Bossard, L.; Guillaumin, M.; and Gool, L.~V. 2014.
\newblock Food-101 - Mining Discriminative Components with Random Forests.
\newblock In Fleet, D.~J.; Pajdla, T.; Schiele, B.; and Tuytelaars, T., eds.,
  \emph{ECCV}, volume 8694 of \emph{Lecture Notes in Computer Science},
  446--461.

\bibitem[{Brown et~al.(2020)Brown, Mann, Ryder, Subbiah, Kaplan, Dhariwal,
  Neelakantan, Shyam, Sastry, Askell, Agarwal, Herbert{-}Voss, Krueger,
  Henighan, Child, Ramesh, Ziegler, Wu, Winter, Hesse, Chen, Sigler, Litwin,
  Gray, Chess, Clark, Berner, McCandlish, Radford, Sutskever, and
  Amodei}]{conf/nips/BrownMRSKDNSSAA20}
Brown, T.~B.; Mann, B.; Ryder, N.; Subbiah, M.; Kaplan, J.; Dhariwal, P.;
  Neelakantan, A.; Shyam, P.; Sastry, G.; Askell, A.; Agarwal, S.;
  Herbert{-}Voss, A.; Krueger, G.; Henighan, T.; Child, R.; Ramesh, A.;
  Ziegler, D.~M.; Wu, J.; Winter, C.; Hesse, C.; Chen, M.; Sigler, E.; Litwin,
  M.; Gray, S.; Chess, B.; Clark, J.; Berner, C.; McCandlish, S.; Radford, A.;
  Sutskever, I.; and Amodei, D. 2020.
\newblock Language Models are Few-Shot Learners.
\newblock In Larochelle, H.; Ranzato, M.; Hadsell, R.; Balcan, M.; and Lin, H.,
  eds., \emph{NeurIPS}.

\bibitem[{Chen et~al.(2015)Chen, Fang, Lin, Vedantam, Gupta, Doll{\'{a}}r, and
  Zitnick}]{journals/corr/ChenFLVGDZ15}
Chen, X.; Fang, H.; Lin, T.; Vedantam, R.; Gupta, S.; Doll{\'{a}}r, P.; and
  Zitnick, C.~L. 2015.
\newblock Microsoft {COCO} Captions: Data Collection and Evaluation Server.
\newblock \emph{CoRR}, abs/1504.00325.

\bibitem[{Chen et~al.(2020)Chen, Li, Yu, Kholy, Ahmed, Gan, Cheng, and
  Liu}]{conf/eccv/ChenLYK0G0020}
Chen, Y.; Li, L.; Yu, L.; Kholy, A.~E.; Ahmed, F.; Gan, Z.; Cheng, Y.; and Liu,
  J. 2020.
\newblock {UNITER:} UNiversal Image-TExt Representation Learning.
\newblock In Vedaldi, A.; Bischof, H.; Brox, T.; and Frahm, J., eds.,
  \emph{ECCV}, volume 12375 of \emph{Lecture Notes in Computer Science},
  104--120.

\bibitem[{Cimpoi et~al.(2014)Cimpoi, Maji, Kokkinos, Mohamed, and
  Vedaldi}]{conf/cvpr/CimpoiMKMV14}
Cimpoi, M.; Maji, S.; Kokkinos, I.; Mohamed, S.; and Vedaldi, A. 2014.
\newblock Describing Textures in the Wild.
\newblock In \emph{CVPR}, 3606--3613.

\bibitem[{Cui et~al.(2021)Cui, Wu, Liu, Yang, and Zhang}]{conf/acl/CuiWLYZ21}
Cui, L.; Wu, Y.; Liu, J.; Yang, S.; and Zhang, Y. 2021.
\newblock Template-Based Named Entity Recognition Using {BART}.
\newblock In Zong, C.; Xia, F.; Li, W.; and Navigli, R., eds., \emph{ACL
  Findings}, volume {ACL/IJCNLP} 2021 of \emph{Findings of {ACL}}, 1835--1845.

\bibitem[{Deng et~al.(2009)Deng, Dong, Socher, Li, Li, and
  Fei{-}Fei}]{conf/cvpr/DengDSLL009}
Deng, J.; Dong, W.; Socher, R.; Li, L.; Li, K.; and Fei{-}Fei, L. 2009.
\newblock ImageNet: {A} large-scale hierarchical image database.
\newblock In \emph{CVPR}, 248--255.

\bibitem[{Devlin et~al.(2019)Devlin, Chang, Lee, and
  Toutanova}]{conf/naacl/DevlinCLT19}
Devlin, J.; Chang, M.; Lee, K.; and Toutanova, K. 2019.
\newblock {BERT:} Pre-training of Deep Bidirectional Transformers for Language
  Understanding.
\newblock In Burstein, J.; Doran, C.; and Solorio, T., eds., \emph{NAACL-HLT},
  4171--4186.

\bibitem[{Dou et~al.(2022)Dou, Xu, Gan, Wang, Wang, Wang, Zhu, Zhang, Yuan,
  Peng, Liu, and Zeng}]{conf/cvpr/DouXGWWWZZYP0022}
Dou, Z.; Xu, Y.; Gan, Z.; Wang, J.; Wang, S.; Wang, L.; Zhu, C.; Zhang, P.;
  Yuan, L.; Peng, N.; Liu, Z.; and Zeng, M. 2022.
\newblock An Empirical Study of Training End-to-End Vision-and-Language
  Transformers.
\newblock In \emph{CVPR}, 18145--18155.

\bibitem[{Fei{-}Fei, Fergus, and Perona(2007)}]{journals/cviu/Fei-FeiFP07}
Fei{-}Fei, L.; Fergus, R.; and Perona, P. 2007.
\newblock Learning generative visual models from few training examples: An
  incremental Bayesian approach tested on 101 object categories.
\newblock \emph{Comput. Vis. Image Underst.}, 106(1): 59--70.

\bibitem[{Gao et~al.(2021)Gao, Geng, Zhang, Ma, Fang, Zhang, Li, and
  Qiao}]{journals/corr/abs-2110-04544}
Gao, P.; Geng, S.; Zhang, R.; Ma, T.; Fang, R.; Zhang, Y.; Li, H.; and Qiao, Y.
  2021.
\newblock CLIP-Adapter: Better Vision-Language Models with Feature Adapters.
\newblock \emph{CoRR}, abs/2110.04544.

\bibitem[{Goyal et~al.(2017)Goyal, Khot, Summers{-}Stay, Batra, and
  Parikh}]{conf/cvpr/GoyalKSBP17}
Goyal, Y.; Khot, T.; Summers{-}Stay, D.; Batra, D.; and Parikh, D. 2017.
\newblock Making the {V} in {VQA} Matter: Elevating the Role of Image
  Understanding in Visual Question Answering.
\newblock In \emph{CVPR}, 6325--6334.

\bibitem[{Guo, Rush, and Kim(2021)}]{conf/acl/GuoRK20}
Guo, D.; Rush, A.~M.; and Kim, Y. 2021.
\newblock Parameter-Efficient Transfer Learning with Diff Pruning.
\newblock In Zong, C.; Xia, F.; Li, W.; and Navigli, R., eds., \emph{ACL},
  4884--4896.

\bibitem[{He et~al.(2022)He, Zhou, Ma, Berg{-}Kirkpatrick, and
  Neubig}]{conf/iclr/HeZMBN22}
He, J.; Zhou, C.; Ma, X.; Berg{-}Kirkpatrick, T.; and Neubig, G. 2022.
\newblock Towards a Unified View of Parameter-Efficient Transfer Learning.
\newblock In \emph{ICLR}.

\bibitem[{Helber et~al.(2019)Helber, Bischke, Dengel, and
  Borth}]{journals/staeors/HelberBDB19}
Helber, P.; Bischke, B.; Dengel, A.; and Borth, D. 2019.
\newblock EuroSAT: {A} Novel Dataset and Deep Learning Benchmark for Land Use
  and Land Cover Classification.
\newblock \emph{{IEEE} J. Sel. Top. Appl. Earth Obs. Remote. Sens.}, 12(7):
  2217--2226.

\bibitem[{Houlsby et~al.(2019)Houlsby, Giurgiu, Jastrzebski, Morrone,
  de~Laroussilhe, Gesmundo, Attariyan, and Gelly}]{conf/icml/HoulsbyGJMLGAG19}
Houlsby, N.; Giurgiu, A.; Jastrzebski, S.; Morrone, B.; de~Laroussilhe, Q.;
  Gesmundo, A.; Attariyan, M.; and Gelly, S. 2019.
\newblock Parameter-Efficient Transfer Learning for {NLP}.
\newblock In Chaudhuri, K.; and Salakhutdinov, R., eds., \emph{ICML}, volume~97
  of \emph{Proceedings of Machine Learning Research}, 2790--2799.

\bibitem[{Hu et~al.(2022)Hu, Shen, Wallis, Allen{-}Zhu, Li, Wang, Wang, and
  Chen}]{conf/iclr/HuSWALWWC22}
Hu, E.~J.; Shen, Y.; Wallis, P.; Allen{-}Zhu, Z.; Li, Y.; Wang, S.; Wang, L.;
  and Chen, W. 2022.
\newblock LoRA: Low-Rank Adaptation of Large Language Models.
\newblock In \emph{ICLR}.

\bibitem[{Huang et~al.(2021)Huang, Zeng, Huang, Liu, Fu, and
  Fu}]{conf/cvpr/HuangZH0FF21}
Huang, Z.; Zeng, Z.; Huang, Y.; Liu, B.; Fu, D.; and Fu, J. 2021.
\newblock Seeing Out of the Box: End-to-End Pre-Training for Vision-Language
  Representation Learning.
\newblock In \emph{CVPR}, 12976--12985.

\bibitem[{Huang et~al.(2020)Huang, Zeng, Liu, Fu, and
  Fu}]{journals/corr/abs-2004-00849}
Huang, Z.; Zeng, Z.; Liu, B.; Fu, D.; and Fu, J. 2020.
\newblock Pixel-BERT: Aligning Image Pixels with Text by Deep Multi-Modal
  Transformers.
\newblock \emph{CoRR}, abs/2004.00849.

\bibitem[{Jia et~al.(2022)Jia, Tang, Chen, Cardie, Belongie, Hariharan, and
  Lim}]{journals/corr/abs-2203-12119}
Jia, M.; Tang, L.; Chen, B.; Cardie, C.; Belongie, S.~J.; Hariharan, B.; and
  Lim, S. 2022.
\newblock Visual Prompt Tuning.
\newblock In \emph{ECCV}, volume 13693, 709--727.

\bibitem[{Kim, Son, and Kim(2021)}]{conf/icml/KimSK21}
Kim, W.; Son, B.; and Kim, I. 2021.
\newblock ViLT: Vision-and-Language Transformer Without Convolution or Region
  Supervision.
\newblock In Meila, M.; and Zhang, T., eds., \emph{ICML}, volume 139 of
  \emph{Proceedings of Machine Learning Research}, 5583--5594.

\bibitem[{Krause et~al.(2013)Krause, Stark, Deng, and
  Fei{-}Fei}]{conf/iccvw/Krause0DF13}
Krause, J.; Stark, M.; Deng, J.; and Fei{-}Fei, L. 2013.
\newblock 3D Object Representations for Fine-Grained Categorization.
\newblock In \emph{ICCV}, 554--561.

\bibitem[{Lan et~al.(2020)Lan, Chen, Goodman, Gimpel, Sharma, and
  Soricut}]{conf/iclr/LanCGGSS20}
Lan, Z.; Chen, M.; Goodman, S.; Gimpel, K.; Sharma, P.; and Soricut, R. 2020.
\newblock {ALBERT:} {A} Lite {BERT} for Self-supervised Learning of Language
  Representations.
\newblock In \emph{ICLR}.

\bibitem[{Lester, Al{-}Rfou, and Constant(2021)}]{conf/emnlp/LesterAC21}
Lester, B.; Al{-}Rfou, R.; and Constant, N. 2021.
\newblock The Power of Scale for Parameter-Efficient Prompt Tuning.
\newblock In Moens, M.; Huang, X.; Specia, L.; and Yih, S.~W., eds.,
  \emph{EMNLP}, 3045--3059.

\bibitem[{Lewis et~al.(2020)Lewis, Liu, Goyal, Ghazvininejad, Mohamed, Levy,
  Stoyanov, and Zettlemoyer}]{conf/acl/LewisLGGMLSZ20}
Lewis, M.; Liu, Y.; Goyal, N.; Ghazvininejad, M.; Mohamed, A.; Levy, O.;
  Stoyanov, V.; and Zettlemoyer, L. 2020.
\newblock {BART:} Denoising Sequence-to-Sequence Pre-training for Natural
  Language Generation, Translation, and Comprehension.
\newblock In Jurafsky, D.; Chai, J.; Schluter, N.; and Tetreault, J.~R., eds.,
  \emph{ACL}, 7871--7880.

\bibitem[{Li et~al.(2018)Li, Farkhoor, Liu, and Yosinski}]{conf/iclr/LiFLY18}
Li, C.; Farkhoor, H.; Liu, R.; and Yosinski, J. 2018.
\newblock Measuring the Intrinsic Dimension of Objective Landscapes.
\newblock In \emph{ICLR}.

\bibitem[{Li et~al.(2019)Li, Yatskar, Yin, Hsieh, and
  Chang}]{journals/corr/abs-1908-03557}
Li, L.~H.; Yatskar, M.; Yin, D.; Hsieh, C.; and Chang, K. 2019.
\newblock VisualBERT: {A} Simple and Performant Baseline for Vision and
  Language.
\newblock \emph{CoRR}, abs/1908.03557.

\bibitem[{Li and Liang(2021)}]{conf/acl/LiL20}
Li, X.~L.; and Liang, P. 2021.
\newblock Prefix-Tuning: Optimizing Continuous Prompts for Generation.
\newblock In Zong, C.; Xia, F.; Li, W.; and Navigli, R., eds., \emph{ACL},
  4582--4597.

\bibitem[{Liu et~al.(2021)Liu, Zheng, Du, Ding, Qian, Yang, and
  Tang}]{journals/corr/abs-2103-10385}
Liu, X.; Zheng, Y.; Du, Z.; Ding, M.; Qian, Y.; Yang, Z.; and Tang, J. 2021.
\newblock {GPT} Understands, Too.
\newblock \emph{CoRR}, abs/2103.10385.

\bibitem[{Liu et~al.(2019)Liu, Ott, Goyal, Du, Joshi, Chen, Levy, Lewis,
  Zettlemoyer, and Stoyanov}]{journals/corr/abs-1907-11692}
Liu, Y.; Ott, M.; Goyal, N.; Du, J.; Joshi, M.; Chen, D.; Levy, O.; Lewis, M.;
  Zettlemoyer, L.; and Stoyanov, V. 2019.
\newblock RoBERTa: {A} Robustly Optimized {BERT} Pretraining Approach.
\newblock \emph{CoRR}, abs/1907.11692.

\bibitem[{Mahabadi, Henderson, and Ruder(2021)}]{conf/nips/MahabadiHR21}
Mahabadi, R.~K.; Henderson, J.; and Ruder, S. 2021.
\newblock Compacter: Efficient Low-Rank Hypercomplex Adapter Layers.
\newblock In Ranzato, M.; Beygelzimer, A.; Dauphin, Y.~N.; Liang, P.; and
  Vaughan, J.~W., eds., \emph{NeurIPS}, 1022--1035.

\bibitem[{Mahabadi et~al.(2021)Mahabadi, Ruder, Dehghani, and
  Henderson}]{conf/acl/MahabadiR0H20}
Mahabadi, R.~K.; Ruder, S.; Dehghani, M.; and Henderson, J. 2021.
\newblock Parameter-efficient Multi-task Fine-tuning for Transformers via
  Shared Hypernetworks.
\newblock In Zong, C.; Xia, F.; Li, W.; and Navigli, R., eds., \emph{ACL},
  565--576.

\bibitem[{Maji et~al.(2013)Maji, Rahtu, Kannala, Blaschko, and
  Vedaldi}]{journals/corr/MajiRKBV13}
Maji, S.; Rahtu, E.; Kannala, J.; Blaschko, M.~B.; and Vedaldi, A. 2013.
\newblock Fine-Grained Visual Classification of Aircraft.
\newblock \emph{CoRR}, abs/1306.5151.

\bibitem[{Mao et~al.(2022)Mao, Mathias, Hou, Almahairi, Ma, Han, Yih, and
  Khabsa}]{conf/acl/MaoMHAM0YK22}
Mao, Y.; Mathias, L.; Hou, R.; Almahairi, A.; Ma, H.; Han, J.; Yih, S.; and
  Khabsa, M. 2022.
\newblock UniPELT: {A} Unified Framework for Parameter-Efficient Language Model
  Tuning.
\newblock In Muresan, S.; Nakov, P.; and Villavicencio, A., eds., \emph{ACL},
  6253--6264.

\bibitem[{Nilsback and Zisserman(2008)}]{conf/icvgip/NilsbackZ08}
Nilsback, M.; and Zisserman, A. 2008.
\newblock Automated Flower Classification over a Large Number of Classes.
\newblock In \emph{ICVGIP}, 722--729.

\bibitem[{Parkhi et~al.(2012)Parkhi, Vedaldi, Zisserman, and
  Jawahar}]{conf/cvpr/ParkhiVZJ12}
Parkhi, O.~M.; Vedaldi, A.; Zisserman, A.; and Jawahar, C.~V. 2012.
\newblock Cats and dogs.
\newblock In \emph{CVPR}, 3498--3505.

\bibitem[{Petroni et~al.(2019)Petroni, Rockt{\"{a}}schel, Riedel, Lewis,
  Bakhtin, Wu, and Miller}]{conf/emnlp/PetroniRRLBWM19}
Petroni, F.; Rockt{\"{a}}schel, T.; Riedel, S.; Lewis, P. S.~H.; Bakhtin, A.;
  Wu, Y.; and Miller, A.~H. 2019.
\newblock Language Models as Knowledge Bases?
\newblock In Inui, K.; Jiang, J.; Ng, V.; and Wan, X., eds., \emph{ACL},
  2463--2473.

\bibitem[{Plummer et~al.(2017)Plummer, Wang, Cervantes, Caicedo, Hockenmaier,
  and Lazebnik}]{journals/ijcv/PlummerWCCHL17}
Plummer, B.~A.; Wang, L.; Cervantes, C.~M.; Caicedo, J.~C.; Hockenmaier, J.;
  and Lazebnik, S. 2017.
\newblock Flickr30k Entities: Collecting Region-to-Phrase Correspondences for
  Richer Image-to-Sentence Models.
\newblock \emph{IJCV}, 123(1): 74--93.

\bibitem[{Radford et~al.(2021)Radford, Kim, Hallacy, Ramesh, Goh, Agarwal,
  Sastry, Askell, Mishkin, Clark, Krueger, and
  Sutskever}]{conf/icml/RadfordKHRGASAM21}
Radford, A.; Kim, J.~W.; Hallacy, C.; Ramesh, A.; Goh, G.; Agarwal, S.; Sastry,
  G.; Askell, A.; Mishkin, P.; Clark, J.; Krueger, G.; and Sutskever, I. 2021.
\newblock Learning Transferable Visual Models From Natural Language
  Supervision.
\newblock In Meila, M.; and Zhang, T., eds., \emph{ICML}, volume 139 of
  \emph{Proceedings of Machine Learning Research}, 8748--8763.

\bibitem[{Raffel et~al.(2020)Raffel, Shazeer, Roberts, Lee, Narang, Matena,
  Zhou, Li, and Liu}]{journals/jmlr/RaffelSRLNMZLL20}
Raffel, C.; Shazeer, N.; Roberts, A.; Lee, K.; Narang, S.; Matena, M.; Zhou,
  Y.; Li, W.; and Liu, P.~J. 2020.
\newblock Exploring the Limits of Transfer Learning with a Unified Text-to-Text
  Transformer.
\newblock \emph{JMLR}, 21: 140:1--140:67.

\bibitem[{Ren, Kiros, and Zemel(2015)}]{ren2015exploring}
Ren, M.; Kiros, R.; and Zemel, R. 2015.
\newblock Exploring models and data for image question answering.
\newblock \emph{NeurIPS}, 28.

\bibitem[{Ren et~al.(2015)Ren, He, Girshick, and Sun}]{conf/nips/RenHGS15}
Ren, S.; He, K.; Girshick, R.~B.; and Sun, J. 2015.
\newblock Faster {R-CNN:} Towards Real-Time Object Detection with Region
  Proposal Networks.
\newblock In Cortes, C.; Lawrence, N.~D.; Lee, D.~D.; Sugiyama, M.; and
  Garnett, R., eds., \emph{NeurIPS}, 91--99.

\bibitem[{Rombach et~al.(2022)Rombach, Blattmann, Lorenz, Esser, and
  Ommer}]{conf/cvpr/RombachBLEO22}
Rombach, R.; Blattmann, A.; Lorenz, D.; Esser, P.; and Ommer, B. 2022.
\newblock High-Resolution Image Synthesis with Latent Diffusion Models.
\newblock In \emph{CVPR}, 10674--10685.

\bibitem[{Ruiz et~al.(2022)Ruiz, Li, Jampani, Pritch, Rubinstein, and
  Aberman}]{ruiz2022dreambooth}
Ruiz, N.; Li, Y.; Jampani, V.; Pritch, Y.; Rubinstein, M.; and Aberman, K.
  2022.
\newblock DreamBooth: Fine Tuning Text-to-image Diffusion Models for
  Subject-Driven Generation.

\bibitem[{Soomro, Zamir, and Shah(2012)}]{journals/corr/abs-1212-0402}
Soomro, K.; Zamir, A.~R.; and Shah, M. 2012.
\newblock {UCF101:} {A} Dataset of 101 Human Actions Classes From Videos in The
  Wild.
\newblock \emph{CoRR}, abs/1212.0402.

\bibitem[{Su et~al.(2020)Su, Zhu, Cao, Li, Lu, Wei, and
  Dai}]{conf/iclr/SuZCLLWD20}
Su, W.; Zhu, X.; Cao, Y.; Li, B.; Lu, L.; Wei, F.; and Dai, J. 2020.
\newblock {VL-BERT:} Pre-training of Generic Visual-Linguistic Representations.
\newblock In \emph{ICLR}.

\bibitem[{Suhr et~al.(2019)Suhr, Zhou, Zhang, Zhang, Bai, and
  Artzi}]{conf/acl/SuhrZZZBA19}
Suhr, A.; Zhou, S.; Zhang, A.; Zhang, I.; Bai, H.; and Artzi, Y. 2019.
\newblock A Corpus for Reasoning about Natural Language Grounded in
  Photographs.
\newblock In Korhonen, A.; Traum, D.~R.; and M{\`{a}}rquez, L., eds.,
  \emph{ACL}, 6418--6428.

\bibitem[{Sung, Cho, and Bansal(2022)}]{conf/cvpr/Sung0B22}
Sung, Y.; Cho, J.; and Bansal, M. 2022.
\newblock {VL-ADAPTER:} Parameter-Efficient Transfer Learning for
  Vision-and-Language Tasks.
\newblock In \emph{CVPR}, 5217--5227.

\bibitem[{Sung, Nair, and Raffel(2021)}]{conf/nips/SungNR21}
Sung, Y.; Nair, V.; and Raffel, C. 2021.
\newblock Training Neural Networks with Fixed Sparse Masks.
\newblock In Ranzato, M.; Beygelzimer, A.; Dauphin, Y.~N.; Liang, P.; and
  Vaughan, J.~W., eds., \emph{NeurIPS}, 24193--24205.

\bibitem[{Vaswani et~al.(2017)Vaswani, Shazeer, Parmar, Uszkoreit, Jones,
  Gomez, Kaiser, and Polosukhin}]{conf/nips/VaswaniSPUJGKP17}
Vaswani, A.; Shazeer, N.; Parmar, N.; Uszkoreit, J.; Jones, L.; Gomez, A.~N.;
  Kaiser, L.; and Polosukhin, I. 2017.
\newblock Attention is All you Need.
\newblock In Guyon, I.; von Luxburg, U.; Bengio, S.; Wallach, H.~M.; Fergus,
  R.; Vishwanathan, S. V.~N.; and Garnett, R., eds., \emph{NeurIPS},
  5998--6008.

\bibitem[{Xiao et~al.(2010)Xiao, Hays, Ehinger, Oliva, and
  Torralba}]{conf/cvpr/XiaoHEOT10}
Xiao, J.; Hays, J.; Ehinger, K.~A.; Oliva, A.; and Torralba, A. 2010.
\newblock {SUN} database: Large-scale scene recognition from abbey to zoo.
\newblock In \emph{CVPR}, 3485--3492.

\bibitem[{Zhang et~al.(2020)Zhang, Sax, Zamir, Guibas, and
  Malik}]{conf/eccv/ZhangSZGM20}
Zhang, J.~O.; Sax, A.; Zamir, A.; Guibas, L.~J.; and Malik, J. 2020.
\newblock Side-Tuning: {A} Baseline for Network Adaptation via Additive Side
  Networks.
\newblock In Vedaldi, A.; Bischof, H.; Brox, T.; and Frahm, J., eds.,
  \emph{ECCV}, volume 12348 of \emph{Lecture Notes in Computer Science},
  698--714.

\bibitem[{Zhang et~al.(2021)Zhang, Fang, Zhang, Gao, Li, Dai, Qiao, and
  Li}]{journals/corr/abs-2111-03930}
Zhang, R.; Fang, R.; Zhang, W.; Gao, P.; Li, K.; Dai, J.; Qiao, Y.; and Li, H.
  2021.
\newblock Tip-Adapter: Training-free CLIP-Adapter for Better Vision-Language
  Modeling.
\newblock \emph{CoRR}, abs/2111.03930.

\bibitem[{Zhou et~al.(2022{\natexlab{a}})Zhou, Yang, Loy, and
  Liu}]{conf/cvpr/ZhouYL022}
Zhou, K.; Yang, J.; Loy, C.~C.; and Liu, Z. 2022{\natexlab{a}}.
\newblock Conditional Prompt Learning for Vision-Language Models.
\newblock In \emph{CVPR}, 16795--16804.

\bibitem[{Zhou et~al.(2022{\natexlab{b}})Zhou, Yang, Loy, and
  Liu}]{journals/ijcv/ZhouYLL22}
Zhou, K.; Yang, J.; Loy, C.~C.; and Liu, Z. 2022{\natexlab{b}}.
\newblock Learning to Prompt for Vision-Language Models.
\newblock \emph{IJCV}, 130(9): 2337--2348.

\bibitem[{Zhou et~al.(2020)Zhou, Ji, Sun, Luo, Hong, Su, Ding, and
  Shao}]{conf/mm/ZhouJSLHSD020}
Zhou, Y.; Ji, R.; Sun, X.; Luo, G.; Hong, X.; Su, J.; Ding, X.; and Shao, L.
  2020.
\newblock K-armed Bandit based Multi-Modal Network Architecture Search for
  Visual Question Answering.
\newblock In Chen, C.~W.; Cucchiara, R.; Hua, X.; Qi, G.; Ricci, E.; Zhang, Z.;
  and Zimmermann, R., eds., \emph{ACM MM}, 1245--1254.

\end{thebibliography}

\end{document}